\documentclass[conference]{IEEEtran}
\IEEEoverridecommandlockouts
\usepackage{cite}
\usepackage{comment}
\usepackage{amsmath,amssymb,amsfonts}
\usepackage{algorithmic}
\usepackage{graphicx}
\usepackage{textcomp}
\usepackage{xcolor}
\usepackage[normalem]{ulem}

\def\BibTeX{{\rm B\kern-.05em{\sc i\kern-.025em b}\kern-.08em
    T\kern-.1667em\lower.7ex\hbox{E}\kern-.125emX}}
\usepackage[symbol]{footmisc}

\newcommand{\blue}[1]{#1}

\usepackage{tikz}

\newcommand\copyrighttext{%
  \footnotesize “This work has been submitted to the IEEE for possible publication. \\
  Copyright may be transferred without notice, after which this version may no longer be accessible.”}

\newcommand\copyrightnotice{%
    \begin{tikzpicture}[remember picture,overlay]
        \node[anchor=south,yshift=10pt] at (current page.south) {\fbox{\parbox{\dimexpr\textwidth-\fboxsep-\fboxrule\relax}{\copyrighttext}}};
    \end{tikzpicture}%
}

\begin{document}
\copyrightnotice


\title{\huge Hardware-aware training of models with synaptic delays for digital event-driven neuromorphic processors}


\author{
Alberto Patiño-Saucedo\textsuperscript{a},
Roy Meijer\textsuperscript{b},
Amirreza Yousefzadeh\textsuperscript{c, d},
Manil-Dev Gomony\textsuperscript{b},  \\
Federico Corradi\textsuperscript{b},

Paul Detteter\textsuperscript{c},

Laura Garrido-Regife\textsuperscript{a},
Bernabé Linares-Barranco\textsuperscript{a},
and Manolis Sifalakis\textsuperscript{c} \\
\textsuperscript{a} Instituto de Microelectrónica de Sevilla, CSIC/Universidad de Sevilla, Seville, Spain,\\
\textsuperscript{b} Eindhoven University of Technology, Dept. Electrical Engineering, Eindhoven, Netherlands,\\
\textsuperscript{c} IMEC-Netherlands, Eindhoven, Netherlands, \\
\textsuperscript{d} University of Twente, Twente, Netherlands\\
Corresponding author email: manolis.sifalakis@imec.nl\\
}

\maketitle

\begin{abstract}

Configurable synaptic delays are a basic feature in many neuromorphic neural network hardware accelerators. However, they have been rarely used in model implementations, despite their promising impact on performance and efficiency in tasks that exhibit complex (temporal) dynamics, as it has been unclear how to optimize them. In this work, we propose a framework to train and deploy, in digital neuromorphic hardware, highly performing spiking neural network models (SNNs) where apart from the synaptic weights, the per-synapse delays are also co-optimized. Leveraging spike-based back-propagation-through-time, the training accounts for both platform constraints, such as synaptic weight precision and the total number of parameters per core, as a function of the network size. In addition, a delay pruning technique is used to reduce memory footprint with a low cost in performance.
We evaluate trained models in two neuromorphic digital hardware platforms: Intel’s Loihi and Imec's Seneca. Loihi offers synaptic delay support using the so-called Ring-Buffer hardware structure. Seneca does not provide native hardware support for synaptic delays. A second contribution of this paper is therefore a novel area- and memory-efficient hardware structure for acceleration of synaptic delays, which we have integrated in Seneca. 
The evaluated benchmark involves several models for solving the SHD (Spiking Heidelberg Digits) classification task, where minimal accuracy degradation during the transition from software to hardware is demonstrated. To our knowledge, this is the first work showcasing how to train and deploy hardware-aware models parameterized with synaptic delays, on multicore neuromorphic hardware accelerators.

\end{abstract}


\section{Introduction}
\noindent Axonal and dendritic synaptic delays are known to be subject to learning and play a time-warping role in the information propagation in biological neural networks~\cite{stoelzel2017}. Therefore, 
drawing inspiration from neuroscience, many neuromorphic neural network accelerators offer primitives and constructs to facilitate synaptic delays~\cite{akopyan2015truenorth, tang2023seneca, furber2014spinnaker, davies2018loihi, starzyk2020}.

However, up until recently, parameterization of neural networks models with synaptic delays had been a largely unexplored territory (for SNNs and ANNs alike), with only a small corpus of literature historically dedicated in the topic~\cite{cohen1997, waibel1989, santos2017, day1993, duro1999, bone2005, shrestha2016, lea2017, vandenoord2016}.

Nevertheless, recently, a few alternative approaches have started emerging for training such \textit{delay models} using back-propagation~\cite{patino2023empirical, zhang2020, hammouamri2023learning, wang2019, sun2023adaptive}, offering a fresh coverage of the topic with promising results. A shared observation that seems to be confirmed in all of these recent works is that models parameterized with synaptic delays achieve competitive, and often superior, performance than other models (both spiking or non-spiking), with on-average smaller model sizes. In addition in~\cite{patino2023empirical} it was also observed that such models additionally tend to have sparser activity, which makes them more memory and energy efficient, despite the memory overhead in neuromorphic accelerators for implementing the delay structures.

Motivated by the study in~\cite{patino2023empirical}, which defended the energy and memory efficiency of delay models executing on neuromorphic accelerators based on estimations, in this paper we conduct actual experiments with two digital hardware platforms (Intel’s Loihi~\cite{davies2018loihi} and Imec’s Seneca~\cite{tang2023seneca}), in order to assess various aspects of the efficiency of delay models. One key question regards the fidelity of delay models trained offline and deployed for inference on neural accelerators. A second question regards the actual energy/power, latency, and memory efficiency of deploying delay models on hardware. The main contributions of this paper are

\begin{itemize}
    \item a hardware-aware training framework generating (synaptic) delay models suited for hardware acceleration
    \item \blue{a newly proposed hardware structure for accelerating synaptic delays in digital neuromorphic processors, with amortizable memory cost that scales with model density and not network depth or size}
    \item a first-of-its-kind benchmark of (synaptic) delay models executed on neuromorphic hardware accelerators
    \item a validation of the credibility of the methodology presented in\cite{tang2023benchmark} for algorithm-hardware benchmarking
\end{itemize}

\section{Related Work}
A canonical formalization of delays, often seen in the literature has been to parameterize synapses with an additional learnable delay variable, which can be learned with back-propagation~\cite{day1993,duro1999,shrestha2018slayer,wang2019, bone2005}, local Hebbian-like learning rules~\cite{zhang2020}, annealing algorithms~\cite{cohen1997}, etc. An alternative approach featured in TDNNs~\cite{waibel1989,vandenoord2016,Gregor14,lea2017,peddinti2015}, and recently for SNNs~\cite{hammouamri2023learning} involves mapping delays in the spatial domain and train them with autoregressive models as temporal convolution kernels (of fixed receptive fields).

Digital neuromorphic accelerator platforms have typically implemented support structures for synaptic delays using one of two methods: \textit{Ring Buffer}~\cite{furber2014spinnaker, davies2018loihi, morrison2005advancing} or \textit{Shared Delay Queue}\cite{akopyan2015truenorth,tang2023seneca}. A ring buffer is a special type of circular queue where currents with different delays accumulate in separate slots of the queue. When using the ring buffer, the maximum possible delay in the system will be limited to the size of the buffer, and the set of possible delays is linearly distributed i.e., the temporal stride is constant. In this method, there is one ring buffer per neuron; therefore, the memory overhead scales with the number of neurons.
By contrast, a delay queue is a shared structure across neurons, lending to more area/memory efficiency when spike activity is sparse. The axon delay is encoded directly in each spike packet containing a few extra bits to indicate the amount of delay. The destination neuro-synaptic core has several queues, each corresponding to a specific delay amount, connected in a cascade. In this scheme, the number of queues scales with the number of possible delays and not the maximum delay. The size of each queue increases if the queue applies more delay on the spikes.

In \cite{tang2023benchmark} a useful hardware-algorithm benchmarking methodology was introduced, which was adopted in~\cite{patino2023empirical} for producing estimates of the memory and energy cost of deploying synaptic delay models on a neuromorphic accelerator. In this work, we follow this methodology to generate similar reference estimates for the networks that we deployed and contrast them with the actual measurements that we performed.

\section{Methods}
\label{sec:methods}

\subsection{Delay Model Description}
\label{subsec:delay_model}
In this work, we use a multilayer Spiking Neural Network (SNN) based on Leaky Integrate-and-Fire (LIF) neurons. Their dynamic depends on their internal state, known as the membrane potential, $u$, and their time-dependent input, $I$. Also, it considers leaky integration that depends on a time constant, ${\tau}$. This model permits computational efficiency for hardware implementation and, at the same time, reproduces biological processes. In a discrete-time realization of a LIF neuron, the membrane potential updates as follows:

\begin{equation}
    u_k = u_{k-1}e^{-\frac{1}{\tau}} (1-\theta_{k-1}) + I_{k-1}
\end{equation}

\begin{equation}
\theta_{k} = \begin{cases}
1 & {u}_{k} \geq u_{th} \\ 
0 & \text{otherwise}
\end{cases}
\end{equation}

\noindent where ${\theta}$ denotes a function to generate activations or spikes whenever the membrane potential reaches a threshold, $u_{th}$. 

\begin{figure}
    \centering
    \includegraphics[width=6.5cm]{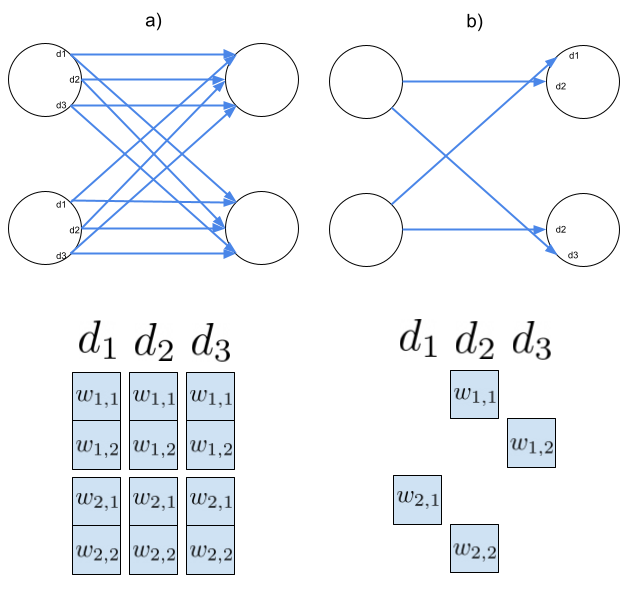}
    \caption{Weight-delay representation before (a) and after (b) prune delay-synapses.}
    \label{fig:Comp_DSNN_RSNN}
\end{figure}

To incorporate synaptic delays in a conventional feed-forward SNN, we create multiple time-delayed projections or synapses for every pre-synaptic/post-synaptic neuron pair. Consequently, the firing of a neuron at a specific time depends on both its current state and a subset of past activations from neurons in the pre-synaptic layer, connected through direct projections. The input of a neuron implementing the suggested framework for synaptic delays is: 

\begin{equation}
I_k= \sum_{d \in D} \sum_{i=1}^{N} w_{i,k}^{(d)} \theta_{i}^{(d)}
\end{equation}

\noindent where  $D=\{0, s, 2s, ... d_{max}\}$ is the set of delays chosen for a given task, which is defined with the maximum delay, $d_{max}$, and the stride, $s$. The set of weights $\{w_{i,j}\}$ in a classic SNN is replaced by $\{w_{i,j}^{(d)}\}$, since the group of weights can take different values for each delay (see Fig.\ref{fig:Comp_DSNN_RSNN}(a)). This increases flexibility in the model by allowing to control the total number of parameters.

\subsection{Training Framework}
\label{subsec:training_framework}
We train models that incorporate synaptic delays using the following strategy, which is compatible with vanilla training frameworks used to train SNNs and RNNs today (i.e. no special framework extensions), such as back-propagation in our case.

The idea is to express the (temporal) parameterization of delays as a spatial parameterization of synaptic weights, such that delay training is effected by merely optimizing for synaptic weights. We start with a set of parallel synapses per pair of pre- and post-synaptic neurons, each associated with a delayed output from the pre-synaptic neuron (using a predetermined range of delays and stride, see Fig.\ref{fig:Comp_DSNN_RSNN}(a)). We optimize the model as usual and prune all delay-synapses that end up with small weights. We then fine-tune the model with only the remaining synapses (see Fig.\ref{fig:Comp_DSNN_RSNN}(b)). We may introduce new synapses to replace the pruned ones, with incrementally higher delay resolution in localized sub-regions of the initial delay range, and repeat the process. As a result, different neurons end up with different fan-in delay inputs. The resulting models are topologically feed-forward, consistently shallower with fewer parameters than their recurrent-connectivity counterparts, and exhibit state-of-art performance (observed in in~\cite{patino2023empirical} and confirmed in other related literature \cite{hammouamri2023learning, sun2022axonal, wang2019}). Their simpler structure renders them attractive for resource-efficient deployment on neuromorphic accelerators.

Note that this training strategy for the delays allows us a priori to take into account hardware constraints such as the maximum delay supported in the platform (e.g. 60 timesteps in Loihi) as well as the delay memory constraints (stride and pruning level may be amortized so that the resulting spike activity does not exceed the available memory).

Another appealing consequence of the simplicity of this strategy is that it is also compatible with on-device local learning rules such as STDP\cite{barranco2011stdp,diehl2015stdp}, as it reduces the goal of learning delays to one of adapting weights.

We train our SNN synaptic delay capable models with spatio-temporal back-propagation (STBP)~\cite{wu2018stbp}, a back-propagation through time (BPTT) variant for SNNs. 
To account for the discontinuity of the membrane potential, we employed a surrogate gradient function~\cite{neftci2019surrogate} with a fast sigmoid function as in ~\cite{zenke2021remarkable}.

\subsection{Hardware Model Deployment and Experimental Setup}

\subsubsection{Network Models and Dataset - SHD}
\label{subsec:dataset}
Using the delay model and training framework that was introduced in the previous sections, we trained three network models for the Spiking Heidelberg Digits (SHD) benchmark\cite{zenke2022shd}: 700-48-48-20, 700-32-32-20, and 700-24-24-20. In each network, synaptic delays are employed between the hidden layers and between the second hidden and output layers. We did not consider synaptic delays from input to the first hidden layer, as the input layer usually has many neurons and is responsible for a large portion of the synaptic parameters. Table~\ref{tb:train_configs} details the trained network topologies, training configurations, initial delay configuration per hidden layer, as well as the final number of delay levels retained after the iterative synapse pruning, and the reference accuracy achieved for each model on the test-set. The initial delay configurations are provided as max delay level and stride, which for example (60, 2) would imply that we utilize synapses with delay levels \{0,2,4 .. 58\}.  
\blue{Finally, for the 700-48-48-20 model, we also explored a variation involving \emph{axonal-only delay pruning}. This means that instead of removing individual synapses, we removed entire groups of synapses per axon until only 15 delay axons remained (700-48-48-20 Ax). The goal here was to understand the trade-off in hardware resource savings between these two pruning strategies.}
Fig.~\ref{fig:methodology} illustrates the steps in the training pipeline.

\begin{table}
    \centering
    \caption{The models trained with synaptic delays for the experiments}
    \label{tb:train_configs}
    \begin{tabular}{|c|c|c|c|}
        \hline
        \textbf{Parameters} & \textbf{Model 1}  & \textbf{Model 2}  & \textbf{Model 3} \\ 
        \hline 
        Topology            & 700-48-48-20      & 700-32-32-20      &  700-24-24-20 \\
        Num timesteps       & 64                & 64                &  64 \\
        Max delay, stride   & 60,2              & 60,2              &  60,2 \\ 
        \# Delays after prune & 15              & 15                 & 15 \\ 
        \# Parameters       &  82.6K           & 47.4K             & 32.6K \\ 
        \hline
        Ref. Accuracy       & 87\%              & 82\%              & 83\% \\
        \hline 
    \end{tabular} 
\end{table}

\subsubsection{Model deployment on Seneca and Loihi}
\label{subsec:model_mapping}
\begin{figure}
    \centering
    \includegraphics[width=5.5cm]{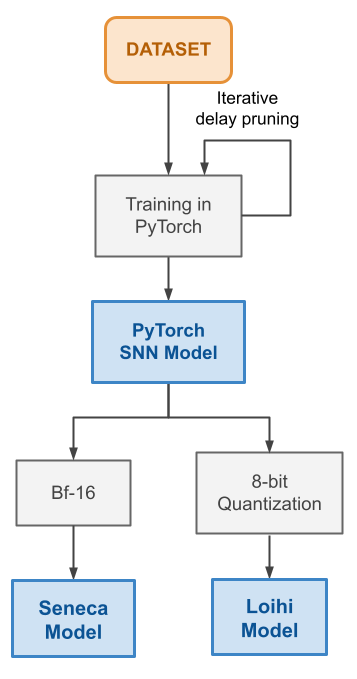}
    \caption{Delay model training pipeline.}
    \label{fig:methodology}
\end{figure}

We used PyTorch to train the SNN models with synaptic delays. After training, we perform hardware-aware fine-tuning with weight quantization and type conversion to produce models that can execute on the neuromorphic accelerators that we used. In the case of Imec's Seneca, we quantized the models to 16-bit Brain-float, and in the case of Intel's Loihi to 8-bit integer (see Fig.\ref{fig:methodology}).

On Seneca, we map the networks to 3 cores (one layer per core, and each core is served by one shared delay queue).
\blue{
In Loihi, we could fit each model in one core, but we also considered a mapping on three cores for the comparison with Seneca, in terms of energy and latency (as this accounts also the overheads for inter-core routing).}

Loihi implements delays with a structure similar to ring-buffers, while Seneca uses a variation of shared delay queues, which are introduced in the next section. 
\blue{Tests on Seneca were done with the acceleration of synaptic delays using the shared delay queue as well as without, whereby delays were emulated in C within the RISC-V controller in each core. This way we can have a quantitative assessment of the cost of using hw-support for synaptic delays.}

\subsection{Memory Efficient Synaptic Delay Acceleration (Seneca)}
\label{subsec:syndelay_seneca}

\subsubsection{Shared Circular Delay Queue Architecture}
\label{subsec:scdq_arch}
\begin{figure}
\centering
\includegraphics[width=0.48\textwidth,trim=15 15 15 15,clip]{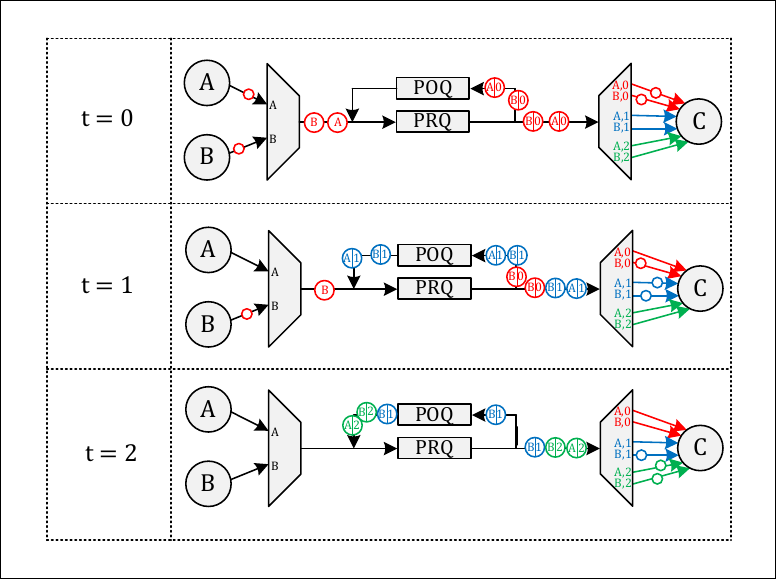}
\caption{
\blue{\small An example of the event flow over three timesteps in a two-layer SNN with synaptic delays. The maximum delay is 2, and the Shared Circular Delay Queue is positioned between the two layers. In timestep $t=0$, neurons $A$ and $B$ spike, and neuron $C$ receives spikes from neurons $A$ and $B$ with a delay value of 0. In timestep $t=1$, neuron $B$ spikes, and neuron $C$ receives a spike from neuron $A$ with a delay value of 0, from neuron $B$ with a delay value of 0, and from neuron $B$ with a delay value of 1. In timestep $t=2$, no neuron spikes, and neuron $C$ receives a spike from neuron $A$ with a delay value of 2, from neuron $B$ with a delay value of 2, and from neuron $B$ with a delay value of 1.}
}
\label{fig:example_3_timesteps}
\end{figure}

\blue{
The proposed synaptic delay structure draws inspiration from the shared queue model in ~\cite{akopyan2015truenorth} and incorporates the concept of ``double-buffering''. Instead of a linear chain of FIFOs (one per delay level), it simplifies the design to only two FIFOs: the \emph{pre-processing queue} (PRQ) and the \emph{post-processing queue} (POQ). These two FIFOs are interconnected in a circular arrangement (see Figure \ref{fig:example_3_timesteps}), hence the name ``Shared Circular Delay Queue'' (SCDQ).
}

\blue{
There is an SCDQ module at every compute core of a neural network accelerator. Conceptually it is positioned between a presynaptic and a postsynaptic layer, with the input of the SCDQ queue facing the presynaptic layer and the output of the SCDQ facing the postsynaptic layer, and is shared across all synapses between neuron pairs. If multiple layers can be mapped in the same compute core, then an SCDQ can also be shared across layers. As a result, the memory complexity is not a function of the number of layers in a model, but rather a function of the number of compute cores in use (and the aggregate activation traffic as we will explain).
}

\blue{
When an event packet (AER) is received at the SCDQ input, a delay counter is added to it, which counts the maximum number of algorithmic timesteps by which the event should be delayed. The event then traverses the PRQ and arrives at the output if delivery to some postsynaptic neuron is expected at the current algorithmic timestep. It will (\emph{also}) be pushed to the POQ if it is expected that a delayed copy will be delivered at a future timestep. At the POQ, the delay counter of events is decremented by one timestep. At some point, a special event packet signaling the \emph{end of the algorithmic timestep} is intercepted at the output and will trigger a buffer-swap between POQ and PRQ. When this happens, the events previously held in POQ will be propagated through PRQ towards the output (if their delay counter has reached a due delay value for delivery in this new timestep), and/or will be pushed back again to the POQ for further delaying and delivery at a future timestep. When the delay counter of an event expires, this event will no longer be pushed to the POQ and will only be forwarded to the output. Figure \ref{fig:example_3_timesteps} exemplifies such a flow of events through the SCDQ across two layers of an SNN network with delays of three timesteps.
}

\blue{
It is important to note that by contrast to its ancestral architecture~\cite{akopyan2015truenorth}, SCDQ makes it possible for an event to ``orbit around'' and exit the queue at several timesteps. This way it can support event delays not only per axon but also per axon-dendrite pair (i.e. per individual synapse), which we will discuss further in section \ref{sec:results}.
}

\begin{figure}
\centering
\includegraphics[width=0.48\textwidth,trim=22 18 22 23,clip]{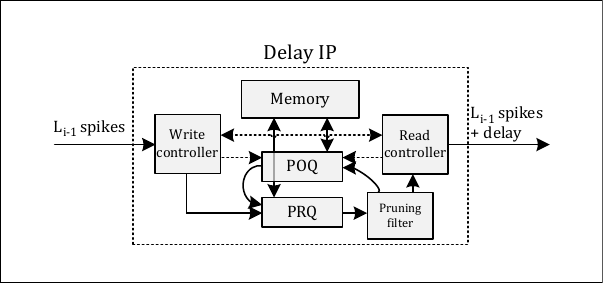}
\caption{
\blue{\small A mapping on the Seneca platform of a three-layer SNN with 700 wide input samples, 48 neurons in both hidden layers, 20 neurons in the output layer, and 60 delays between the hidden layers}
}
\label{fig:delay_ip_accelerator}
\end{figure}

\blue{
SCDC consists of six submodules shown in the Delay IP block diagram of Figure \ref{fig:delay_ip_accelerator}: the \emph{write controller} and \emph{read controller} which get triggered by external transactions, the \emph{PRQ} and \emph{POQ}, the \emph{pruning filter}, and \emph{memory}. The \emph{write controller} is responsible for reading the events from the input, writing them to the PRQ, and adding to them an initialized delay counter value. Additionally, it is responsible for swapping the PRQ and POQ buffers when a special \emph{end-of-timestep} (EOT) event is intercepted at the output. The \emph{read controller} is responsible for reading events from the PRQ and writing them to the output and the POQ while decreasing the delay counter. It will also signal the write controller to start the EOT sequence when there is an EOT event at the output. The PRQ and POQ are FIFO abstractions around memory buffers available in the memory module and thereby implement the actual data-path through the SCDQ.
}

\subsubsection{Zero-Skipping Delay-Forwarding}
\label{subsec:scdq_pruning}
\blue{
One observes in Figure \ref{fig:Comp_DSNN_RSNN} that the network topology with multiple delay synapses can become quite intricate. The number of synapses increases linearly with the maximum number of delay steps, $D$, which we aim to accommodate. In practical scenarios, weight training-quantization-pruning, and the potential use of suitable regularization removes a substantial number of synapses, keeping only those that effectively contribute to the model performance. This results in a much more compact model for practical deployment on an accelerator.
}

\blue{
In a zero-skipping capable accelerator, weights are typically stored in sparse matrix format, and a zero weight will not trigger memory IO. A zero activation will similarly skip fetching weights from memory and computing the partial MAC, thereby in both cases saving energy and likely improving latency.
}

\blue{
When accelerating synaptic delays with SCDQ this zero-skipping decision is not taken on event generation, but rather after an event exits the queue and is about to be broadcast to the postsynaptic neurons. This is because even if an event activation is not useful at the current timestep, it may still be in future timesteps.
This situation may unnecessarily occupy FIFO memory, consume power for repeated lookups in weight-memory, and affect latency in SCDQ, when some or all of the postsynaptic neuron connections have zero weights at certain timesteps.
}

\begin{figure}[h]
    \centering
    \begin{minipage}{0.3\textwidth}
        \includegraphics[width=\linewidth,trim=15 15 15 15,clip]{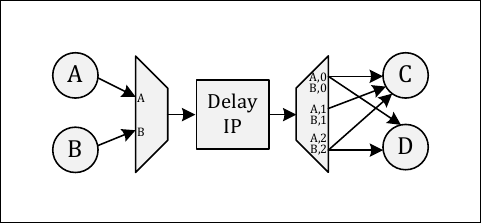}
    \end{minipage}\hfill
    \begin{minipage}{0.15\textwidth}
        \[
        WVU =
        \begin{bmatrix}
        1 & 1 & 0\\
        0 & 0 & 1
        \end{bmatrix}
        \]
    \end{minipage}
    \caption{
    \blue{\small Example of $WVU$ for a network where delayed axons $A,2$, $B,0$ and $B,1$ are skipped.}
    }
    \label{fig:wvu_example}
\end{figure}

\blue{
To limit this overhead a \emph{zero-skiping delay-forwarding} capability is implemented by means of the \emph{pruning filter} of SCDQ. A binary matrix called $WVU$ (which stands for ``weight value useful'') is maintained locally, where every element represents a delayed axon. $WVU$ has two dimensions: dimension $I$ representing every presynaptic neuron, and dimension $D$ representing every delay level. If $WVU_{i,d} = 1$, this means that for the current delay value $d$ and the presynaptic neuron address $i$, at least one of the weights is a non-zero value.
If $WVU_{i,d} = 0$, this means that for all $i$ and $d$, the weights are 0
(which implies that event delivery can be skipped at the current timestep). Using this method, the SCDQ can determine for every event in the PRQ if it is useful for the postsynaptic layer (or if it should be ignored). Figure \ref{fig:wvu_example} shows an example of a network with some pruned delayed axons and the corresponding  $WVU$ matrix.
}

\blue{
To determine when an event should be removed from the delay queue (i.e. every next timestep of the event, $WVU_{i,d}$ will be 0), a row of $WVU$, which corresponds to some presynaptic neuron $i$, can be treated as a binary number. If we count the number of leading zeros (\texttt{clz(x)}) of this binary number, we know for which delay counter value $d$ the event does not have to move from the PRQ to the POQ anymore. For example, in Figure \ref{fig:wvu_example}, \texttt{clz($WVU_{A}$) = 1} and \texttt{clz($WVU_{B}$) = 0}, which means events from neuron $A$ have to stay for 2 timesteps (can be removed at delay counter 1) and events from neuron $B$ do not have to stay for 3 timesteps (can be removed at delay counter 0).
}

\section{Results}
\label{sec:results}

For the tests on Loihi, we had available a Kapoho Bay USB stick with dual Loihi chip 128 cores each, and remote access to a Nahuku board (32 chips) via the Intel Neuromorphic Research Community (INRC). 
\blue{For power measurements on Loihi, we used Intel's cloud Loihi server farm, as it is currently the only way to perform power measurements of neurosynaptic cores. We repeated all the tests one hundred times and averaged the results to account for variance in power measurements.
}

\blue{Imec's Seneca is not taped out in ASIC yet, which is what allowed us to integrate the delay acceleration circuit and evaluate it. We therefore provide results from circuit-level simulations and estimate power estimations based on a netlist provided by Imec, in a 22 nm process from Global Foundry. Energy and area measurements are acquired using the Cadence Joules Xcelium at 500MHz and Genus tools, respectively. The Joules tool provides accuracy within 15\% of the sign-off power. The Genus tool is used to estimate the area. Latency, memory, and fidelity measurements are obtained from the hardware test bench simulations.
}

\subsection{Fidelity of delay models on neuromorphic hardware}
\begin{figure}
    \centering
    \includegraphics[width=8.5cm]{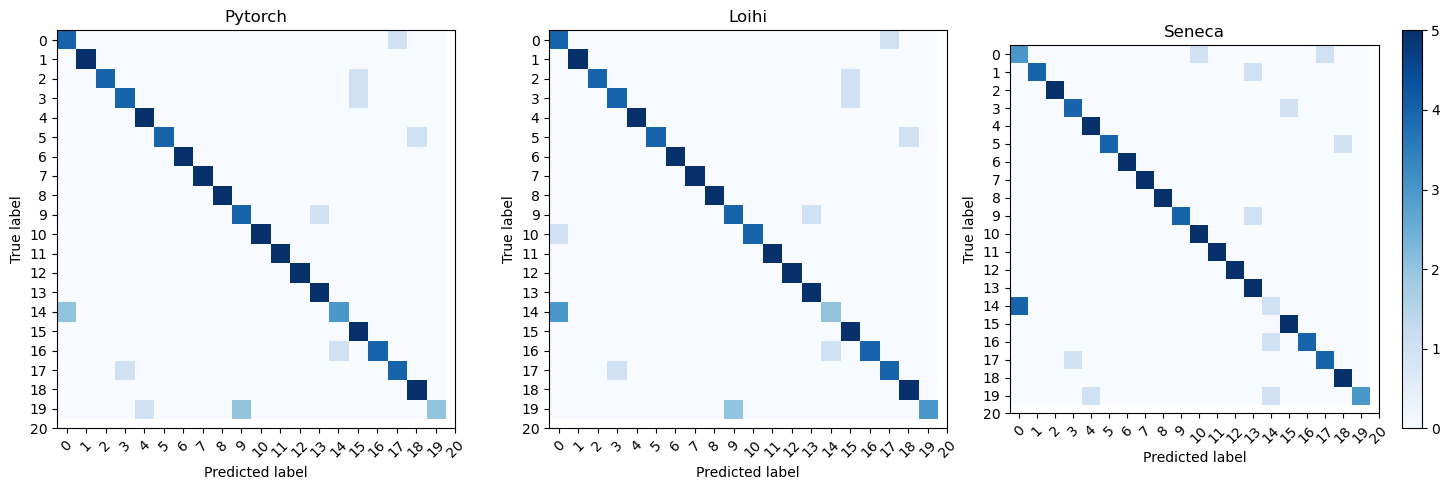}
    \caption{Confusion matrices for PyTorch trained ``mother-model'' and Loihi/Seneca run models for the 700-48-48-20 network.}
    \label{fig:class_fidelity}
\end{figure}

The first aspect that we were concerned with verifying, was the fidelity in terms of model performance, between the PyTorch-trained model (``mother model'') and the fine-tuned models for the two neuromorphic accelerators. Table~\ref{tb:accuracy_results} summarises the rather positive observations. Both quantized models running on Loihi and Seneca were very close in behavior and final task performance to the ``mother model''. There is about 1\% deviation in task accuracy between inference in PyTorch and inference in the neuromorphic accelerators. Very importantly, the predictions made by the hardware models were very consistent with those of the ``mother model''. This was well above 90\% consistency across all classes, for all models. A more detailed breakdown per class for the 700-48-48-20 network is shown in Fig~\ref{fig:class_fidelity}. 
\blue{
The accuracy of 48-48-20 Ax (axonal delay pruning) is decreased (not the fidelity between software and hardware though) because of the axon grouping constraint for pruning that penalizes axons containing the 15 smallest weights (but which could contain some large weights too. However, the goal is to show later the improvements in energy and latency, when pruning entire axons.
}

\begin{figure}
    \centering
    \includegraphics[width=9cm]{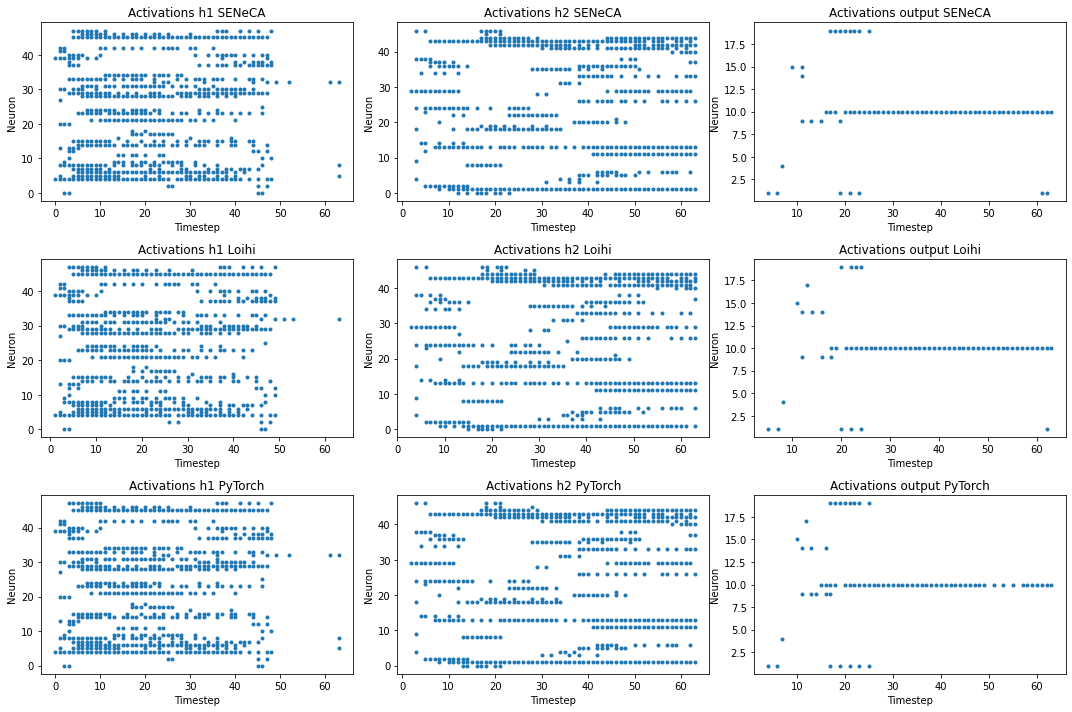}
    \caption{Comparison of spike traces between pytorch ``mother-model'' and the one executed on Seneca and Loihi for SHD testset sample X.}
    \label{fig:seneca_spike_trace}
\end{figure}

\begin{figure}
    \centering
    \includegraphics[width=9cm]{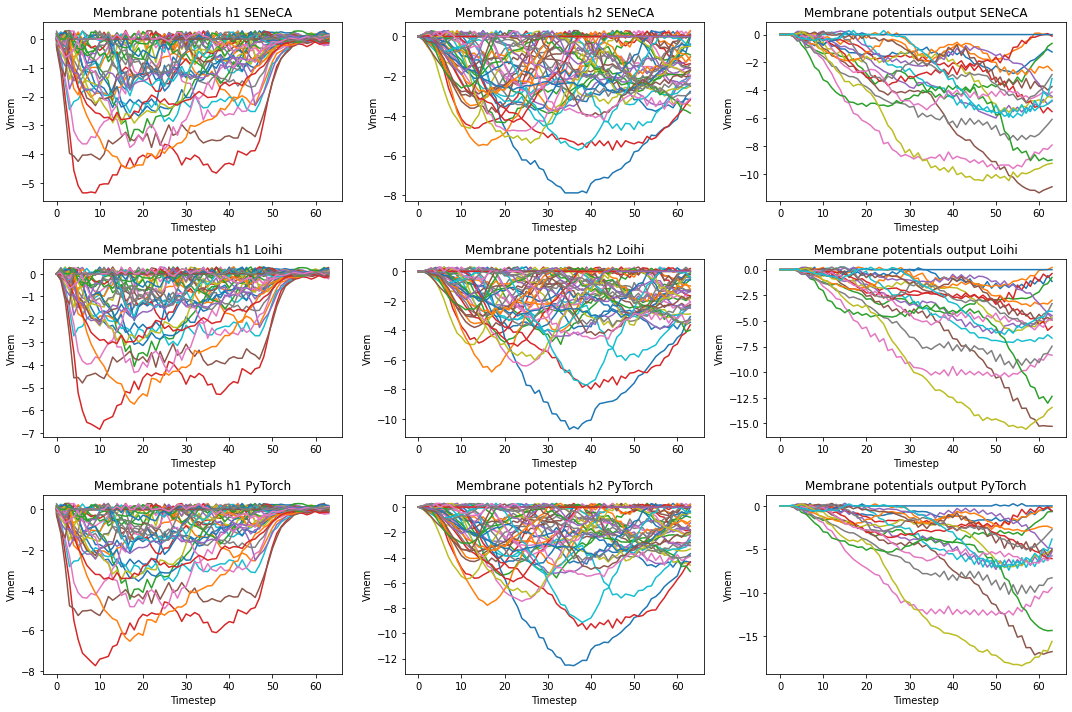}
    \caption{Comparison of Vmem trace between pytorch ``mother-model'' and the one executed on Seneca and Loihi for SHD testset sample X.}
    \label{fig:seneca_vmem_trace}
\end{figure}

Last and even more important is the fact that the spike traces and membrane evolution traces inside the network are also very consistent between ``mother model'' and the hardware executing models. This can be verified both in the numbers reported Table~\ref{tb:accuracy_results} for the entire test-set and the three models, and also visually in Fig~\ref{fig:seneca_spike_trace} and \ref{fig:seneca_vmem_trace} for one example data point. It means that the software model can provide us with very reliable information about the anticipated energy per inference and memory use on the neuromorphic platform without even running the hardware model (since these traces are a proxy for the number of synaptic operations).
 
\begin{table}[]
    \centering
    \caption{Fidelity of Loihi/Seneca run models against the PyTorch trained ``mother-model''}
    \renewcommand{\arraystretch}{1.1}    
    \label{tb:accuracy_results}
\begin{tabular}{|ccccccc|}
\hline
\multicolumn{1}{|c|}{\textbf{Model}}        & \multicolumn{2}{c|}{\textbf{Pytorch}}         & \multicolumn{2}{c|}{\textbf{Loihi}}           & \multicolumn{2}{c|}{\textbf{Seneca}} \\ \hline
\multicolumn{7}{|c|}{Model accuracy}                                                                                                           \\ \hline
\multicolumn{1}{|c|}{700-48-48-20} & \multicolumn{2}{c|}{87\%}            & \multicolumn{2}{c|}{87\%}            & \multicolumn{2}{c|}{86\%}   \\
\multicolumn{1}{|c|}{700-32-32-20} & \multicolumn{2}{c|}{82\%}            & \multicolumn{2}{c|}{83\%}            & \multicolumn{2}{c|}{81\%}   \\
\multicolumn{1}{|c|}{700-24-24-20} & \multicolumn{2}{c|}{83\%}            & \multicolumn{2}{c|}{81\%}            & \multicolumn{2}{c|}{82\%}   \\
\multicolumn{1}{|c|}{700-48-48-20 Ax} & \multicolumn{2}{c|}{76\%}            & \multicolumn{2}{c|}{75\%}            & \multicolumn{2}{c|}{76\%}\\ \hline
\multicolumn{7}{|c|}{Prediction consistency against Pytorch}                                                                                   \\ \hline
\multicolumn{1}{|c|}{700-48-48-20} & \multicolumn{2}{c|}{100\%}           & \multicolumn{2}{c|}{97\%}            & \multicolumn{2}{c|}{93\%}   \\
\multicolumn{1}{|c|}{700-32-32-20} & \multicolumn{2}{c|}{100\%}           & \multicolumn{2}{c|}{95\%}            & \multicolumn{2}{c|}{97\%}   \\
\multicolumn{1}{|c|}{700-24-24-20} & \multicolumn{2}{c|}{100\%}           & \multicolumn{2}{c|}{93\%}            & \multicolumn{2}{c|}{98\%}   \\
\multicolumn{1}{|c|}{700-48-48-20 Ax} & \multicolumn{2}{c|}{100\%}           & \multicolumn{2}{c|}{92\%}            & \multicolumn{2}{c|}{94\%}   \\ \hline
\multicolumn{7}{|c|}{Avg spike count per inference per layer}                                                                                  \\ \hline
\multicolumn{1}{|c|}{}             & h1     & \multicolumn{1}{c|}{h2}     & h1     & \multicolumn{1}{c|}{h2}     & h1           & h2           \\
\multicolumn{1}{|c|}{700-48-48-20} & 545.32 & \multicolumn{1}{c|}{534.55} & 524.39 & \multicolumn{1}{c|}{531.14} & 544.71       & 534.25       \\
\multicolumn{1}{|c|}{700-32-32-20} & 400.99 & \multicolumn{1}{c|}{384.61} & 398.33 & \multicolumn{1}{c|}{365.15} & 401.18       & 384.85       \\
\multicolumn{1}{|c|}{700-24-24-20} & 273.86 & \multicolumn{1}{c|}{311.19} & 272.48 & \multicolumn{1}{c|}{300.75} & 271.99       & 307.9        \\ \hline
\end{tabular}

\end{table}

\begin{figure}
    \centering
    \includegraphics[width=8.5cm]{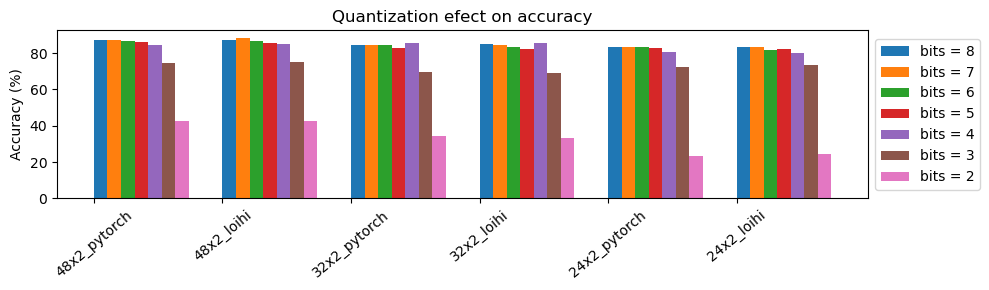}
    \caption{Effect of weight quantization on final accuracy. All test-set (2264 samples) are considered.}
    \label{fig:loihi_quant}
\end{figure}

\subsection{Power, Energy, Latency, Memory measurements}

\blue{
In Table \ref{tb:hw_metrics} we report (average) hardware measurements per inference in Loihi, in Seneca with Delay IP, and in Seneca without Delay IP (that is delays implemented in software on the RISC-V controller). The power consumption reported in the table includes only the total power of the neurosynaptic cores, excluding the I/O and peripherals. Energy is calculated using dynamic power consumption, to exclude the high 74.4 mW static power consumption of the Lakemont x86 management cores of Loihi. The static power consumption of Seneca is 462 $\mu W$. Comparing the middle and right columns justifies the circuit-based implementation of SCDQ in Seneca as it improves energy efficiency and latency by a significant amount. The energy efficiency increases between $3.5 \times$ (700-48-48-20) and $3.1 \times$ (700-24-24-20). Similarly, latency decreases between $4.3 \times$ (700-48-48-20) and $3.5 \times$ (700-24-24-20). In other words, the benefits increase with the total activation volume (seen in larger networks). For the axonal pruned network (700-48-48-20 Ax), both energy efficiency and latency improve by approximately $1.4 \times$ compared to the per individual synapse pruned network (700-48-48-20). This is because the activation sparsity is higher for 700-48-48-20 Ax than for 700-48-48-20.
}

\blue{
As a proxy comparison for the two different hardware implementations of synaptic delays, namely Ring Buffer in Loihi, versus SCDQ (improved shared queue) in Seneca, the energy efficiency of Seneca (SCDQ) appears approximately $1.2-1.9 \times$ lower, but note that this involves 1 core in Loihi versus 3 cores on Seneca.
But the gap is closing as the network size increases, which reveals the better scalability of the SCDQ and/or the costly contribution of the RISC-V controller in Seneca: the larger the network, the more compute work gets allocated to the vector-pipeline of neuro co-processors (NPEs), compared to the work the RISC-V controller has to do.
}

\blue{
What is also very interesting is the effect of the accelerated synaptic delays with SCDQs on the inference latency. On Seneca, both with the hardware implementation and the RISC-V implementation of SCDQ, the latency decreases with the network size (due to fewer activations), while it, surprisingly, remains constant on Loihi's Ring Buffers. And for the hardware accelerated implementation, it is consistently less than $0.5 \times$ that of Loihi, despite the larger number of cores used and routing between them (i.e., longer data path).
}

\blue{
To make the comparison more equitable we performed an additional experiment where the 700-48-48-20 network was distributed across three cores in Loihi too. In this case (Table \ref{tb:hw_metrics_3c}) energy consumption increases by 35\% and memory use by a factor of $1.15 \times$. 
}

\begin{figure}
    \centering
    \includegraphics[width=7cm]{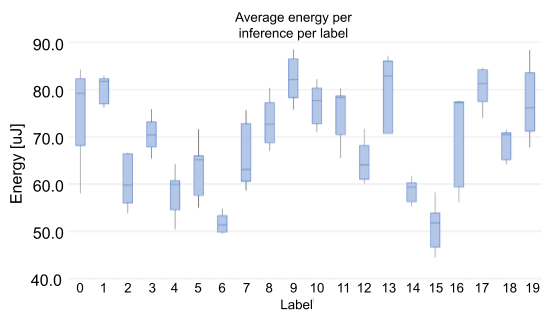}
    \caption{Energy consumption per inference by class on Seneca.}
    \label{fig:seneca_energy_class}
\end{figure}

\blue{
In Seneca, the shared queues (SCDQs) contribute only 2\% - 3\% to its energy consumption, with the remaining 97\% - 98\% coming from the cores processing (of which 67\% - 82\% of this energy consumption is related to pre- and post-processing events in the RISC-V controller of each core). Similarly, out of the 3.3Mb memory consumption, RISC-V instructions take up 0.5Mb, while neural processing operations use up 2.6Mb. It's worth noting that the actual energy and memory consumption of neural processing operations (14.6uJ/2.6Mb) are very close to the estimates generated using the methodology introduced in \cite{tang2023benchmark} (14.48uJ/2.2Mb). Fig~\ref{fig:seneca_energy_class}, provides a more detailed breakdown of the energy boxplots for the per-class data points in the test set for Seneca for the network 700-24-24-20.
}


\begin{table}[] 
    \footnotesize
    \centering
    \caption{\small Average hardware measurements per inference, DIP = Delay IP, Energy calculated using dynamic power consumption (on Loihi this does not include the Lakemont cores.)}
    \renewcommand{\arraystretch}{1.2}    
    \label{tb:hw_metrics}
    \begin{tabular}{|c|ccc|}
        \hline
        \textbf{Measurement} & \textbf{Loihi (1c)} & \textbf{Seneca (3c) w/ SCDQ} & \textbf{Seneca (3c)  w/o SCDQ} \\
        \hline
        \multicolumn{4}{|c|}{\textbf{Network: 700-48-48-20}}\\
        \hline 
        \textit{Energy}       & 28.4 $\mu J$     			& 43.6 $\mu J$   	& 152 $\mu J$\\ 
        \textit{Latency}      & 9.1 $ms$					& 4.25 $ms$   		& 18.3 $ms$ 	\\
        \textit{Memory}       & 1.96 $Mb$         			& 3.38 $Mb$   		& 3.34 $Mb$  	\\
        \textit{DIP energy}   & NA             				& 0.98 $\mu J$   	& 0 $\mu J$    	\\
        \hline 
        \multicolumn{4}{|c|}{\textbf{Network: 700-32-32-20}}									\\
        \hline 
        \textit{Energy}       & 18.3 $\mu J$                & 29.5 $\mu J$  	& 92.4 $\mu J$  \\ 
        \textit{Latency}      & 9.1 $ms$                    & 3.34 $ms$   		& 13.6 $ms$ 	\\
        \textit{Memory}       & 1.48 $Mb$         	        & 2.33 $Mb$   		& 2.29 $Mb$  	\\
        \textit{DIP energy}   & NA             			    & 0.77 $\mu J$   	& 0 $\mu J$    	\\
        \hline 
        \multicolumn{4}{|c|}{\textbf{Network: 700-24-24-20}}									\\
        \hline 
        \textit{Energy}       & 12.5 $\mu J$                & 23.8 $\mu J$   	& 73.3 $\mu J$ 	\\ 
        \textit{Latency}      & 9.1 $ms$         	        & 2.81 $ms$   		& 9.81 $ms$ 	\\
        \textit{Memory}       & 1.23 $Mb$         	        & 2.07 $Mb$   		& 2.05 $Mb$  	\\
        \textit{DIP energy}   & NA                          & 0.62 $\mu J$      & 0 $\mu J$    	\\
        \hline 
        \multicolumn{4}{|c|}{\textbf{Network: 700-48-48-20 Ax}}									\\
        \hline 
        \textit{Energy}       & 25.2 $\mu J$                & 31.4 $\mu J$      & NA 			\\ 
        \textit{Latency}      & 10.0 $ms$         	        & 3.03 $ms$   		& NA 			\\
        \textit{Memory}       & 1.96 $Mb$         	        & 3.36 $Mb$   		& NA 			\\
        \textit{DIP energy}   & NA             			    & 0.72 $\mu J$      & NA 			\\
        \hline 
    \end{tabular} 
\end{table}

\begin{table}[] 
    \centering
    \caption{Hardware measurements for executing inference with model 700-48-48-20 using 3 cores on Loihi and Seneca} 
    \renewcommand{\arraystretch}{1.2}    
    \label{tb:hw_metrics_3c}
    \begin{tabular}{|c|cc|}
        \hline
        \textbf{HW measurement}  & \textbf{Loihi (3c)}    & \textbf{Seneca (3c)}\\
        \hline
        \multicolumn{3}{|c|}{\textbf{model: 700-48-48-20}}\\
        \hline 
        Energy per inference    & 38.4 $\mu J$  & 43.6 $\mu J$ \\
        Latency per inference   & 9.1 $ms$      & 4.25 $ms$ \\
        Memory consumption      & 4.2 $Mb$      & 3.38 $Mb$ \\
        Synaptic Delay energy &  NA             & 0.98 $\mu J$ \\
        \hline 
    \end{tabular} 

\end{table}

\subsection{Final accuracy and effect of reduced bit precision}
We report the accuracy on the entire test set (2264 samples) and the effect of quantizing with different bit precisions for the three base models implemented on Loihi. As observed in Fig. \ref{fig:loihi_quant} and consistent with the fidelity results, the degradation in accuracy of the mapping form Pytorch to Loihi is negligible. Interestingly, the effect of reducing the bit precision is, at most less than 3\% percent for up to 4 bits, leading (potentially) to a further reduction of the memory footprint with little cost. The maximum accuracy obtained in Loihi for the 700-48-48-20 model is 89.04\%

\section{Discussion}
\label{sec:discussion}

\blue{
Regarding the cost of supporting synaptic delays in a neural accelerator, there are two sources of overhead. First delay parameterization involves extending the weight matrix structure from 2D to 3D (2D weight matrix for each delay value). It has dimensions $D \times I \times J$, where $D$ is the delay resolution (that is, the number of distinct discrete values), $I$ is the number of pre-synaptic neurons, and $J$ is the number of post-synaptic neurons. Pruning during training can serve to limit or reduce this overhead (particularly when the weight matrix can be stored in compressed form). Pruning an axon removes a set of delay synapses towards post-synaptic layer neurons (reducing the cardinality of one of the dimensions of the weight matrix), and pruning a dendritic branch alone removes an individual synaptic connection between a pre-post neuron pair.
}

\blue{
Second, in most neuromorphic architectures, delays are implemented as FIFO queues or delay lines placed either at every neuron or shared among synaptic connections. For example, in~\cite{davies2018loihi, khan2008spinnaker,morrison2005advancing} a \emph{Ring Buffer} places a circular FIFO queue placed at each postsynaptic neuron where dendritic weights accumulate. Each \emph{slot} of the Ring Buffer accumulates weights for a different timestep. At the end of a timestep, the slot corresponding to the current timestep is added to the membrane potential of the corresponding postsynaptic neuron. Subsequently, the Ring Buffer shifts to the next timestep, and the slot from the current timestep becomes the slot for the maximum delay. The number of delay levels is constrained by the size (number of slots) of the Ring Buffer. With only one Ring Buffer per postynaptic neuron, the total memory overhead is the product of the number of postynaptic neurons $J$ and the number of delay levels $D$, scaling with $\mathcal{O}(J \cdot D)$ \cite{patino2023empirical}.
}

\blue{
On the other hand, a \emph{Shared Queue}~\cite{akopyan2015truenorth}, of which SCQD is an extension, 
shares a group of FIFOs in a linear cascade arrangement across the synapses (presynaptic and postsynaptic neuron pairs of two layers). The number of FIFOs in the cascade equals the maximum number of discrete delay levels $D$. Every event triggered at a presynaptic neuron from an axon with a given delay, enters the Shared Queue at the FIFO corresponding to the number of timesteps it needs to wait before delivery. At the end of a timestep, all events in every FIFO shift to the next FIFO in the cascade belonging to the next relative timestep, and all events in the FIFO belonging to the current timestep are transmitted to the postsynaptic neurons. The total memory usage depends on the minimum activation sparsity of the network (which can be biased through model training). Assuming a minimum activation sparsity of $\alpha$, where $\alpha = 1$ implies that every presynaptic neuron is activated and $\alpha = 0$ implies that none are activated, the total memory overhead is calculated as $\alpha \cdot I \cdot \left(\sum_{d=1}^{D}(D-d)\right) = \frac{1}{2}\cdot \alpha \cdot I \cdot (D^2 + D)$, where $I$ is the number of presynaptic neurons and $D$ is the number of delay levels. This memory overhead scales as $\mathcal{O}(\alpha \cdot I \cdot D^2)$. In this synaptic delay model, only axonal delays are supported and therefore an event enters and exits the Shared Delay Queue only once.
}

\blue{
The SCDQ introduced in this paper and employed in ~\cite{tang2023seneca} is an evolution of the Shared Queue in~\cite{akopyan2015truenorth}. SCDQ has memory overhead of $\alpha \cdot I \cdot (2\cdot D - 1)$, scaling with $\mathcal{O}(\alpha \cdot I \cdot D)$. By reference to TrueNorth's\cite{akopyan2015truenorth} constraints of 16 timesteps of delay, 256 neurons, and for sparsity $\alpha = 1$ (no spasity), TrueNorth's Shared Delay Queue needs queue capacity for $34816$ events, while SCDQ only requires storage for $7936$ events. In the implementation of the SCDQ we used, each event has a bit width of 16 bits. By comparison to a Ring Buffer structure using as a reference Loihi's\cite{davies2018loihi} constraints of 64 delay time steps, 48 neurons, and 8 bits per weight, the additional memory overhead of Loihi is only $24576$ bits, which is less than the $97536$ bits of the Delay IP. However, this comparison assumes no activation sparsity ($\alpha = 1$), and the memory overhead of Ring Buffers does not scale with activation sparsity. If $\alpha \leq 0.25$, the SCDQ has a lower memory overhead than Loihi's Ring Buffers.
SpiNNaker\cite{khan2008spinnaker} also uses Ring Buffers and has a constraint of 16-time steps of delay, 256 neurons, and 16-bit weights per neuron, resulting in a memory overhead of $65536$ bits. Assuming $\alpha = 1$, the SCDQ has a memory overhead of $126976$ bits in this case and when $\alpha \leq 0.5$, SCDQ is more memory efficient than SpiNNaker too.
}

\begin{figure}
\centering
\includegraphics[width=0.40\textwidth,trim=5 5 5 5,clip]{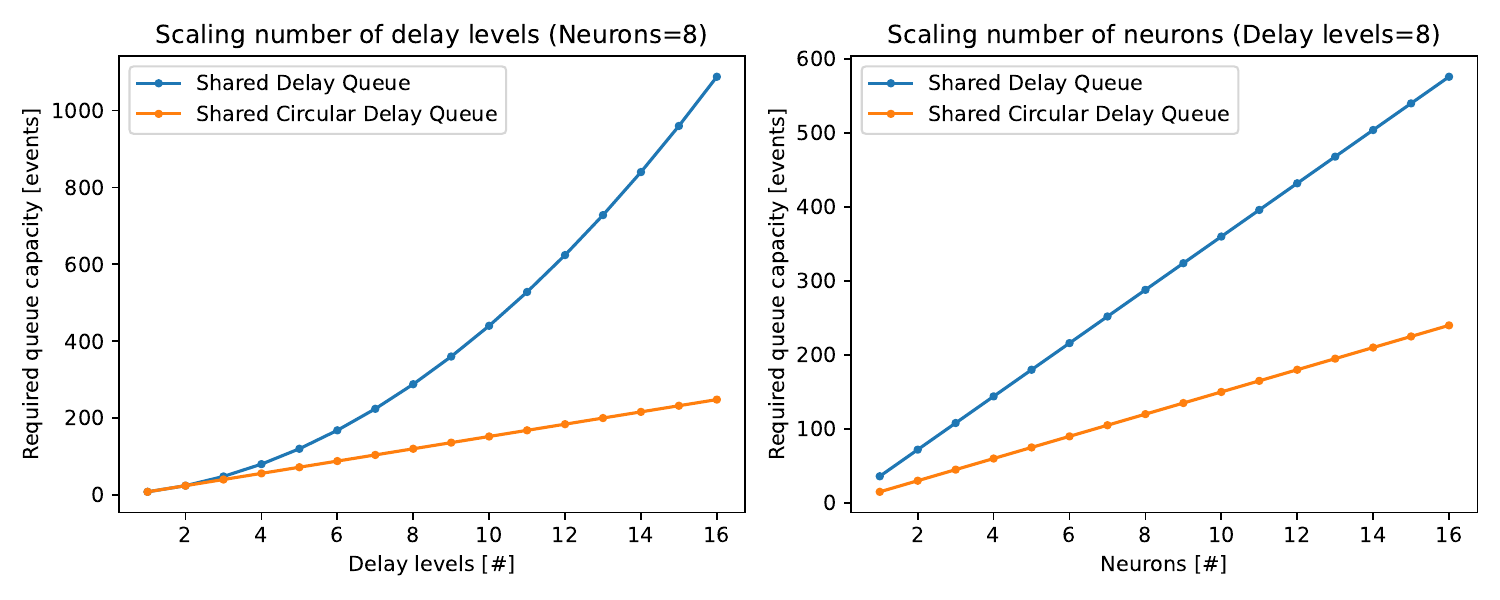}
\caption{\small Required memory for shared delay queue vs shared circular delay queue}
\label{fig:queue_size_scaling}
\end{figure}

\blue{
To illustrate how SCDQ improves upon the \emph{vanilla} Shared Delay Queue structure~\cite{akopyan2015truenorth}, in Figure~\ref{fig:queue_size_scaling}, we provide a worst-case estimate (assuming dense activations and no delay pruning) between two fully connected layers. The figure illustrates the buffer capacity required for both a Shared Delay Queue and SCDQ, as the number of delay levels increases and as the number of postsynaptic neurons increases.
}

\blue{
Finally, considering area overhead, in the initial implementation of SCDQ we used flip-flops. To reduce the total area, flip-flop memory was later replaced with a 2048-word 16-bit SRAM memory, with enough capacity for the observed maximum of 1596 events in the queues. This resulted in an 81\% reduction in total area, with a total of 15463 $\mu m^2$ for the Delay IP (see Table \ref{tb:area}). With this area improvement the area cost of SCDQ reduces from $\frac{81624}{472400} = 17\%$ of the area one Seneca core down to a mere $\frac{15463}{472400} = 3\%$ only.
}

\begin{table}[ht]
    \footnotesize
    \centering
    \setlength{\tabcolsep}{2pt} 
    \caption{\small Area Savings with SRAM}
    \label{tb:area}
    \begin{tabular}{|c|cc|cc|cc|c|}
    \hline
   \textbf{Memory} & \multicolumn{2}{c|}{\textbf{Combinational}} & \multicolumn{2}{c|}{\textbf{Flip-Flop}} & \multicolumn{2}{c|}{\textbf{SRAM}} & \textbf{Total}\\\hline
   \textit{FF}   & 16885 $\mu m^2$ & 21\% & 64739 $\mu m^2$ & 79\% & 0 $\mu m^2$ & 0\% & 81624 $\mu m^2$ \\\hline
   \textit{SRAM} & 2287 $\mu m^2$ & 15\% & 4557 $\mu m^2$ & 29\% & 8618 $\mu m^2$ & 56\% & 15463 $\mu m^2$ \\\hline
   \multicolumn{7}{|c|}{\textit{Area saved}} & 81\%\\\hline
\end{tabular}
\end{table}

\section{Conclusion}
In this paper, \blue{we have proposed} a framework for training and deploying highly performing spiking neural network \blue{models} (SNNs) with synaptic delays on digital neuromorphic hardware. The framework co-optimizes both synaptic weights and delays \blue{and takes hardware platform constraints into account}. \blue{We have also introduced a new hardware data structure, SCDQ, for memory- and area-efficient acceleration of synaptic delays that combines the ideas of ring buffers and shared queues. We evaluated our contributions with} two neuromorphic digital hardware platforms: Intel's Loihi and Imec's Seneca \blue{(in which SCDQ was integrated). Our evaluation with several models of varying sizes trained for the SHD classification task, affirms not only the efficacy and hardware efficiency of synaptic delays but also our ability under the proposed training framework to produce hardware models that perform on par with their software counterparts on digital hardware accelerators.}

This work demonstrates the first successful application of hardware-aware models parameterized with synaptic delays on multi-core neuromorphic hardware accelerators.

\section*{Acknowledgments}
This work was funded by EU Horizon grant 101070679 "NimbeAI" and EU H2020 grant 871501 "Memscales".

\bibliographystyle{IEEEtran}
\bibliography{references}

\end{document}